\documentclass[a4paper, 10pt, conference]{ieeeconf}      
\usepackage{FG2020}
\usepackage{multirow}

\FGfinalcopy 

\IEEEoverridecommandlockouts                              
\usepackage{cite}
\usepackage{graphicx}
\usepackage{booktabs}

\def\FGPaperID{210} 

\title{\LARGE \bf
Estimation of Orofacial Kinematics in Parkinson's Disease: \\ Comparison of 2D and 3D Markerless Systems for Motion Tracking}

\author{\parbox{16cm}{\centering
    {\large Diego L. Guarin$^1$, Aidan Dempster$^1$, Andrea Bandini$^1$,\\ Yana Yunusova$^{1,2,3}$ and Babak Taati$^{1,4,5}$}\\
    {\normalsize
    $^1$ KITE | Toronto Rehabilitation Institute | University Health Network, Toronto, ON, Canada.\\
    $^2$ Department of Speech Language Pathology, University of Toronto, Toronto, ON, Canada.
    \\
    $^3$ Hurvitz Brain Sciences Program, Sunnybrook Research Institute, Toronto, ON, Canada. \\
    $^4$ Institute of Biomaterials and Biomedical Engineering, University of Toronto, Toronto, ON, Canada.\\		 
	$^5$ Department of Computer Science, University of Toronto, Toronto, ON, Canada. }}
    \thanks{DLG is supported by the Michael J. Fox Foundation for Parkinson's Research}
}

\begin{document}

\ifFGfinal
\thispagestyle{empty}
\pagestyle{empty}
\else
\author{Anonymous FG2020 submission\\ Paper ID \FGPaperID \\}
\pagestyle{plain}
\fi
\maketitle

\begin{abstract}
Orofacial deficits are common in people with Parkinson's disease (PD) and their evolution might represent an important biomarker of disease progression. We are developing an automated system for assessment of orofacial function in PD that can be used in-home or in-clinic and can provide useful and objective clinical information that informs disease management. Our current approach relies on color and depth cameras for the estimation of 3D facial movements. However, depth cameras are not commonly available, might be expensive, and require specialized software for control and data processing. The objective of this paper was to evaluate if depth cameras are needed to differentiate between healthy controls and PD patients based on features extracted from orofacial kinematics. Results indicate that 2D features, extracted from color cameras only, are as informative as 3D features, extracted from color and depth cameras, differentiating healthy controls from PD patients. These results pave the way for the development of a universal system for automatic and objective assessment of orofacial function in PD.
\end{abstract}

\section{INTRODUCTION}

Orofacial symptoms are common in Parkinson's disease (PD)~\cite{schneider1986deficits}, and the severity of orofacial motor disorders may be correlated with the severity of PD as measured by clinical scales~\cite{bakke2011orofacial, fereshtehnejad2017evolution}. Thus, there has been great interest in developing techniques to objectively evaluate orofacial movements in PD. Researchers have introduced multiple approaches to measure the kinematic characteristics of the tongue, jaw, and lips, such as position and velocity, during speech production~\cite{ kearney2017sentence, bologna2016altered, ackermann1997kinematic, forrest1995dynamic}. These methods provide objective information on how PD affects normal function and have the potential to improve the clinical management of the disease. However, the clinical utility of these techniques is limited as they depend on expensive and difficult-to-use systems, such as optical motion tracking systems, electromagnetic articulography, and electromyography. 

Bandini \textit{et al.} introduced an alternative approach to estimate orofacial kinematics that uses a color and depth cameras. The technique employs a computer vision model to automatically  estimate the position of facial landmarks -- defining the location of the eyebrows, eyes, nose, mouth, and jawline -- in videos acquired with the color camera. The information provided by the depth camera is subsequently used to reconstruct the 3D \textit{real-word} locations of the facial landmarks~\cite{bandini2016markerless, bandini2017video, bandini2018automatic}. Results demonstrated that this approach was able to differentiate between healthy controls and PD patients based only on the information provided by a few short video segments of subjects performing a task commonly applied during neurological examination~\cite{bandini2017analysis}.

This approach represents an important step towards an objective, automatic, cost-effective, and universally available system for assessment of orofacial deficits in PD. Such a system would be available for in-clinic or in-home use with little or no intervention from movement disorder specialists, and would provide clinically useful information that informs disease management. However, the clinical utility of the system proposed by Bandini \textit{et al.} is limited by the necessity of a depth camera, which might not be available in the clinic or at home. Thus, a truly universally available system should rely only on color cameras, which are already available on mobile phones, tablets, and personal computers. 

To our knowledge, only one work analyzed the ability of 2D and 3D orofacial kinematic features to differentiate between healthy controls and PD patients. Bandini \textit{et al.} analyzed one speech task -- repetition of the syllable /pa/ -- and found that only 3D features were significantly different between healthy controls and PD patients~\cite{bandini2016markerless}. This paper builds upon the work of Bandini \textit{et al.} and analyzes the motion of the lips during execution of speech and non-speech tasks from 8 PD patients and 12 healthy controls. Kinematics parameters were extracted in i) 3D using  a combination of color and depth cameras, and ii) 2D using only a color camera, with the objective of evaluating if 2D information were sufficient for assessing of orofacial function in PD patients and healthy controls.  

\section{Materials and Methods}
\subsection{Participants}
Twenty participants were recruited for this study: eight patients with Parkinson's disease (1 female, age $66.9 \pm 20.5$ years), and twelve age-matched healthy controls (8 female, age $72.9 \pm 12.5$ years). The study was approved by the University of Toronto's Research Ethics Board. Participants signed an informed consent form according to the requirements of the Declaration of Helsinki.   
\subsection{Experimental setup}
Participants were seated in front of an Intel RealSense D400, consisting of a registered depth and color camera pair, with a face-to-camera distance of 30-50~cm~\cite{bandini2016markerless}. A continuous light source was placed adjacent to the camera to provide uniform illumination. Participants were asked to look at the camera and were recorded during the execution of standard neurological assessment tasks. A video composed of color (RGB) and depth information was recorded for each task. Both streams were recorded at $\sim$30 frames per second at VGA resolution (640 x 480 pixels). A total of 80 videos were included in the analysis, 48 from healthy controls, and 32 from patients. An audio file was also recorded for the duration of the procedure; a flashlight with a clicker was used to synchronize audio and video recordings. 
\subsection{Experimental procedure}
Participants were asked to perform a set of speech and non-speech tasks commonly used during neurological evaluation of orofacial performance~\cite{bandini2018automaticALS}. Based on previous results reported in experiments involving people with Parkinson's disease~\cite{bandini2016markerless}, amyotrophic lateral sclerosis~\cite{bandini2018automaticALS}, and stroke~\cite{bandini2018automatic}, four tasks were considered in this study. These task included repetition of the sentence '\textit{Buy Bobby a Puppy}' 5 times at a comfortable rate and loudness (BBP); repetition of the syllable /\textit{pa}/ as fast as possible on a single breath (PA); making a big smile showing teeth 5 times (BIGSMILE); and maintaining a neutral facial expression, eyes open, and mouth closed for 20 s (REST). Participant were encouraged to take breaks between tasks to prevent fatigue.   
\subsection{Pre-processing}
Data pre-processing consisted of three steps applied in sequence: (1) Task segmentation by repetition, (2) face alignment, and (3) reconstruction of 3D information. Pre-processing was performed with a custom scrip written in Python.
\subsubsection{Task segmentation}
All tasks, except REST, were manually segmented into individual repetitions by a trained observer; the observer identified the beginning and end of each repetition using the audio or video recordings. The end results of this procedure were a set of videos, each containing a single repetition of the task. 

\subsubsection{Facial alignment}
The Facial Alignment Network (FAN), an open-source, deep-neural network-based framework for automatic facial detection and localization of facial landmarks~\cite{bulat2017far} was used to localize the face and the position of 68 facial landmarks in each video frame. Landmarks outlined the superior border of the brow, the free margin of the upper and lower eyelids, the nasal midline, the nasal base, the mucosal edge and vermillion-cutaneous junction of the upper and lower lips, and the lower two-thirds of the face~\cite{sagonas2013300}.  
\subsubsection{Reconstruction of 3D information}
Color and depth streams were aligned using the camera intrinsic information. Afterwards, the real world coordinates for each landmark were computed.

\subsection{Orofacial features}
\subsubsection{Orofacial properties}
Feature selection was based in previous studies with PD patients. Features were computed from five mouth properties that describe the mouth length and shape during task execution. The properties correspond to the vertical and horizontal mouth opening, the areas of the left and right side of the mouth, and the total mouth area. These properties were computed based on the position of the estimated 2D $\left([x_p, y_p]\right)$ and 3D $\left([x_w,y_w,z_w]\right)$ landmarks for each video frame as follows:

\begin{itemize}
	\item Vertical mouth opening ($TB$), computed as the euclidean distance between the landmarks localized at the top and bottom vermillion borders at the midline; 
	\item Horizontal mouth opening ($WM$), computed as the euclidean distance between the landmarks localized at the left and right oral commissures;
	\item Left mouth area ($AreaLeft$), computed as the area of a triangle formed by the landmarks localized at the top and bottom vermillion borders at the midline and the left oral commissure;
	\item Right mouth area ($AreaRigh$), computed as the area of a triangle formed by the landmarks localized at the top and bottom vermillion borders at the midline and the right oral commissure; and
	\item Overall mouth area ($Area$), computed as the sum of left and right mouth areas.
\end{itemize}

\begin{table*}
	\caption{Standardized Mean Difference (SMD) between Healthy Controls (HC) and Subjects with Parkinson's Disease (PD).}
	\label{table1}
	\resizebox{\textwidth}{!}
	{
		\begin{tabular}{llccrccr}
			\toprule
			&          &  \multicolumn{3}{c}{3D Landmarks} &  \multicolumn{3}{c}{2D Landmarks} \\
			\cmidrule(lr){3-5} 
			\cmidrule(lr){6-8}
			Task & Feature & HC & PD & SMD & HC & PD &SMD\\
			\midrule
			\midrule
			\multirow{7}{*}{BBP} & ${\Delta}TB$ & $1.7\pm0.9$  & $1.1\pm0.3$ & \textbf{0.90}  &$1.2\pm0.4$ & $0.91\pm0.3$ & \textbf{0.84}\\
			& Max velocity $TB$ (1/s) & $36.2\pm27.5$ & $17.2\pm6.1$ & \textbf{0.86} & $19.2\pm6.2$ & $14.1\pm4.7$ & \textbf{0.89}\\
			& Min velocity $TB$ (1/s) & -$30.7\pm26.1$ & -$16.8\pm5.5$ & 0.67 & -$20.0\pm7.8$ & -$15.7\pm5.6$ & 0.69\\
			& Max acceleration $TB$ (1/s$^2$) & $1836.2\pm1605.1$ & $868.9\pm346.1$ & 0.76 & $1041.1\pm430.9$ & $771.7\pm329.8$ & 0.68\\
			& Min acceleration $TB$ (1/s$^2$) & -$2312\pm2204.5$ & -$880.4\pm365.4$ & 0.75 & -$1032.3\pm383.7$ & -$761.9\pm306.3$ & 0.76 \\
			& ${\Delta}Area$ & $1.7\pm1.1$ & $1.2\pm0.4$ & 0.61 & $1.2\pm0.3$ & $0.9\pm0.3$ &\textbf{ 0.80} \\
			& $CCC$~$Area$ & $0.8\pm0.2$ & $0.7\pm0.2$ & 0.18 & $0.6\pm0.2$ & $0.4\pm0.3$ & 0.65 \\
			\midrule
			\multirow{4}{*}{BIGSMILE} & ${\Delta}WM$ & $0.3\pm0.0$ & $0.2\pm0.1$ & \textbf{ 0.85} & $0.3\pm0.0$ & $0.2\pm0.1$ & \textbf{0.84}\\
			& Min velocity $WM$ (1/s) & -$3.4\pm0.9$ & -$2.8\pm0.8$ & 0.66 & -$3.3\pm0.9$ & -$2.7\pm0.9$ & 0.68 \\
			& ${\Delta}Area$ & $1.7\pm0.7$ & $1.5\pm0.7$ & 0.25 & $1.4\pm0.5$ & $1.0\pm0.4$ & 0.61\\
			& $CCC$~$Area$ & $0.9\pm0.1$ & $0.8\pm0.2$ & 0.55 &  $0.8\pm0.2$ & $0.5\pm0.3$ & \textbf{1.24}\\
			\midrule
		\end{tabular}
	}
	{Bold values indicate a \textit{large} difference between groups (SMD$>$0.8)}
\end{table*}

\subsubsection{Normalization of 2D and 3D orofacial properties}
The mean values of the mouth properties computed during the REST task were used as normalization factors for each subject. Only the middle 5~s segment of the REST task were used to compute the normalization factors. The normalization factors were computed by estimating the mouth properties for every frame during the 5~s window (150 video frames) and extracting mean values. 
\subsubsection{Extraction of 2D and 3D features}
Thirteen orofacial kinematic features were extracted from each repetition of '\textit{Buy Bobby a Puppy}', \textit{/pa/}, and big smile, these included: 

\begin{itemize}
	\item  ${\Delta}TB$, computed as the difference between the maximum and minimum values of $TB$
	\item Max velocity $TB$, computed as maximum value of the first derivative of $TB$
	\item Min velocity $TB$, computed as minimum value of the first derivative of $TB$
	\item Max acceleration $TB$, computed as maximum value of the second derivative of $TB$
	\item Min acceleration $TB$, computed as minimum value of the second derivative of $TB$
	\item  ${\Delta}WM$, computed as the difference between the maximum and minimum values of $WM$
	\item Max velocity $WM$, computed as maximum value of the first derivative of $WM$
	\item Min velocity $WM$, computed as minimum value of the first derivative of $WM$
	\item Max acceleration $WM$, computed as maximum value of the second derivative $WM$
	\item Min acceleration $WM$, computed as minimum value of the second derivative of $WM$
	\item Mean $Area$, computed as the mean value of $Area$
	\item ${\Delta}Area$, computed as the difference between the maximum and minimum values of $Area$ 
	\item $CCC$~$Area$, computed as the concordance correlation coefficient between $AreaLeft$ and $AreaRight$ 
\end{itemize}

The concordance correlation coefficient measures the agreement between two signals~\cite{lawrence1989concordance}, and was used as a measure of symmetry between left and right mouth movements. 

\subsection{Statistical analysis}
Each task was analyzed independently. The ability of each feature to differentiate between healthy controls and PD patients was evaluated using the standardized mean difference (SMD)~\cite{cohen2013statistical}, computed as 
\begin{eqnarray*}
	SMD = \displaystyle\frac{\mu_{1}-\mu_{2}}{\sqrt{\displaystyle\frac{(n_1-1)s_{1}^{2} + (n_2 - 1)s_{2}^{2}}{n_{1}+n_2 -2}}}
\end{eqnarray*}

were $\mu_1$, $\mu_2$, $s_{1}$, $s_{2}$, $n_1$, and $n_2$ are the mean, standard deviation, and number of elements for the feature computed for healthy controls and PD patients respectively. An SMD value lower than 0.5 indicates a \textit{small} difference between groups, an SMD value of 0.5 or larger indicates a \textit{medium} difference between groups, and an SMD value of 0.8 or larger indicates a \textit{large} difference between groups~\cite{cohen2013statistical, sawilowsky2009new}. This analysis was performed independently for features obtained from 3D and 2D landmarks.

\section{Results}

Table~\ref{table1} shows the mean $\pm$ standard deviation of  estimated features as well as the SMD between healthy controls (HC) and Parkinson's disease (PD) patients. The table only shows features with a \textit{medium} ($0.5\leq$ SMD $<0.8$) or \textit{large} (SMD $\geq0.8$) difference between groups when extracted with 3D or 2D landmarks.  

These results show that only features extracted from the tasks BBP and BIGSMILE demonstrated \textit{medium} or \textit{large} difference between healthy controls and PD when estimated with 3D or 2D landmarks. In contrast, features extracted from the speech task PA demonstrated only \textit{small} (SMD $<0.5$) difference between groups and are not presented in the table.

\subsection{3D vs. 2D landmarks}
Table~\ref{table1} shows that features computed from 3D landmarks that demonstrated a \textit{medium} or \textit{large} difference between healthy controls and PD patients showed essentially the same behavior when computed from 2D landmarks. In contrast, area related features demonstrated \textit{medium} or \textit{large} difference between groups when computed from 2D landmarks and only \textit{small} or \textit{medium} difference between groups when computed from 3D landmarks.  

\subsection{Relevant features}
\subsubsection{BBP} Features extracted form 3D and 2D landmarks demonstrating a \textit{large} difference between healthy controls and PD include the mouth vertical range of motion (${\Delta}TB$), and its maximum velocity (Max velocity $TB$).The overall change in the mouth area (${\Delta}Area$) also demonstrated large difference between groups, but only when extracted from 2D landmarks.  
\subsubsection{BIGSMILE}
The only feature extracted form 3D and 2D landmarks demonstrating a \textit{large} difference between healthy controls and PD was the mouth horizontal range of motion (${\Delta}WM$). The concordance correlation coefficient between left and right mouth areas ($CCC$~$Area$) also demonstrated large difference between groups, but only when extracted from 2D landmarks.

\section{DISCUSSION}
A universal system for automatic assessment of PD should be based on readily available technology to reach a large number of patients despite geographical and socioeconomic limitations. Standard color cameras are available in mobile phones, tablet, and personal computers, which are readily accessible in most clinics. Thus, standard color cameras represent an ideal technology over which to develop a system for remote, automated, and objective assessment of PD symptoms. 

Previously developed systems automatically identified orofacial deficits in PD based on information provided by color and depth cameras. Depth cameras are available on commercial grade products such as Microsoft Kinect, Azure Kinect, or Intel RealSense. However, these cameras cost as much as \$400USD, are not commonly available in clinics, and require an accompanying computer or laptop with specialized software for data acquisition and processing. Thus, the objective of this study was to compare orofacial kinematic features estimated during the execution of speech and non-speech tasks from information provided by standard color cameras (2D landmarks), and a combination of color and deep cameras (3D landmarks), and evaluate the ability of these features to distinguish between healthy controls and PD patients. 

Our results demonstrated that 2D landmarks information was at least as successful as 3D landmark information in differentiating between healthy controls and PD patients; this observation was validated by the fact that for 3D features with SMD larger or equal than 0.5, their corresponding 2D counterpart also showed a SMD larger or equal than 0.5. By contrast, we observed that some 2D-based features demonstrated a large difference between healthy controls and PD only when extracted from 2D landmarks. We will study this observation in detail in our future work.

Finally, we observed that the most relevant features in 3D and 2D to distinguish between healthy controls and PD patients were related to the vertical (in BBP) and horizontal (in BIGSMILE) movements of the mouth. As it was expected, results indicate that PD patients have smaller range of motion and move slower than healthy subjects. These are cardinal orofacial symptoms of PD~\cite{fereshtehnejad2017evolution}, that might be a consequence of the rigidity, bradykinesia, and akinesia associated with the disease~\cite{de2006epidemiology}. 

\subsection*{Limitations}
Herein, we used pre-trained models for face detection and for localization of facial landmarks. However, these types of models are well know for providing larger landmarks localization error when applied to elderly subjects and patients with neurological diseases~\cite{taati2019algorithmic,asgarian2019limitations}. In our future work, we will re-train or fine-tune the network to improve its accuracy in our target population.

%
%


\bibliographystyle{IEEEtran}
\bibliography{biblio}

\end{document}